\documentclass[nohyperref]{article}

\usepackage{microtype}
\usepackage{graphicx}
\usepackage{subfigure}
\usepackage{booktabs} % for professional tables
\usepackage{enumitem}
\usepackage{mathptmx}

\usepackage{hyperref}

\usepackage[accepted]{icml2022}

\usepackage{amsmath}
\usepackage{amssymb}
\usepackage{mathtools}
\usepackage{amsthm}
\usepackage{svg}

\usepackage[capitalize,noabbrev]{cleveref}

\theoremstyle{plain}

\theoremstyle{definition}

\theoremstyle{remark}

\usepackage[textsize=tiny]{todonotes}

\icmltitlerunning{Empirical Analysis of the AdaBoost's Error Bound}

\begin{document}

\twocolumn[\icmltitle{Empirical Analysis of the AdaBoost's Error Bound}

% It is OKAY to include author information, even for blind
% submissions: the style file will automatically remove it for you
% unless you've provided the [accepted] option to the icml2022
% package.

% List of affiliations: The first argument should be a (short)
% identifier you will use later to specify author affiliations
% Academic affiliations should list Department, University, City, Region, Country
% Industry affiliations should list Company, City, Region, Country

% You can specify symbols, otherwise they are numbered in order.
% Ideally, you should not use this facility. Affiliations will be numbered
% in order of appearance and this is the preferred way.
\icmlsetsymbol{equal}{*}

\begin{icmlauthorlist}
\icmlauthor{Arman Bolatov}{equal,cs}
\icmlauthor{Kaisar Dauletbek}{equal,math}
\end{icmlauthorlist}

\icmlaffiliation{cs}{Department of Computer Science, Nazarbayev University, Astana, Kazakhstan}
\icmlaffiliation{math}{Department of Mathematics, Nazarbayev University, Astana, Kazakhstan}
%\icmlaffiliation{sch}{School of ZZZ, Institute of WWW, Location, Country}

\icmlcorrespondingauthor{Arman Bolatov}{arman.bolatov@nu.edu.kz}
\icmlcorrespondingauthor{Kaisar Dauletbek}{kaisar.dauletbek@nu.edu.kz}

% You may provide any keywords that you
% find helpful for describing your paper; these are used to populate
% the "keywords" metadata in the PDF but will not be shown in the document
\icmlkeywords{Machine Learning, ICML}

\vskip 0.3in
]

% this must go after the closing bracket ] following \twocolumn[ ...

% This command actually creates the footnote in the first column
% listing the affiliations and the copyright notice.
% The command takes one argument, which is text to display at the start of the footnote.
% The \icmlEqualContribution command is standard text for equal contribution.
% Remove it (just {}) if you do not need this facility.

\printAffiliationsAndNotice{}  % leave blank if no need to mention equal contribution
% \printAffiliationsAndNotice{\icmlEqualContribution} % otherwise use the standard text.

\begin{abstract}
Understanding the accuracy limits of machine learning algorithms is essential for data scientists to properly measure performance so they can continually improve their models' predictive capabilities. This study empirically verified the error bound of the AdaBoost algorithm for both synthetic and real-world data. The results show that the error bound holds up in practice, demonstrating its efficiency and importance to a variety of applications. The corresponding source code is available at \href{https://github.com/armanbolatov/adaboost_error_bound}{github.com/armanbolatov/adaboost\_error\_bound}.
\end{abstract}

\section{Introduction}
\label{submission}
In this report, we aim to present an empirical verification of the AdaBoost algorithm \cite{adaboost}. We are going to do so by first showing the theoretical error bounds along with the necessary conditions. Afterward, we will describe an experimental setup and report on the findings. Finally, we will apply the designed experiments on both synthetic and real-world data to provide empirical verification. The theoretical part of this report is based on the "Foundations of Machine Learning" book \cite{foundations_ml}. All experiments are implemented with the machine learning library `scikit-learn` (version 1.2.1) \cite{scikit-learn} and visualized with the `seaborn` (version 0.12.1) \cite{seaborn} and `matplotlib` (version 3.6.1) \cite{Hunter:2007} libraries in Python (version 3.8.2) \cite{van1995python}.

\subsection{AdaBoost Algorithm}
AdaBoost is a boosting algorithm that is designed to construct a strong PAC-learnable \cite{pac} algorithm by means of combining distinct weak PAC-learnable classifiers (base classifiers). The formal algorithm for the implementation of AdaBoost is presented in Algorithm 1.

\begin{algorithm}[tbh]
   \caption{AdaBoost}
   \label{alg:adaboost}
\begin{algorithmic}
   \STATE {\bfseries Input:} data $S$ = (($x_1$, $y_1$), ..., ($x_m$, $y_m$))
   \FOR{$i$ $\xleftarrow{}$ $1$ $\bold{to}$ $m$}
   \STATE $D_1(i) \xleftarrow{} \frac{1}{m}$
   \ENDFOR
   \FOR{$t$ $\xleftarrow{}$ $1$ $\bold{to}$ $T$}
   \STATE $h_t\xleftarrow{}$ classifier in $H$ with $\epsilon_t = P_{i\mathtt{\sim}D_t} [h_t(x_i)\neq y_i]$
   \STATE $\alpha_t\xleftarrow{} \frac{1}{2} \log{\frac{1-\epsilon_t}{\epsilon_t}}$
   \STATE $Z_t\xleftarrow{} 2[\epsilon_t(1-\epsilon_t)]^\frac{1}{2}$ %\vartriangleright $normalization$ $factor$
   
   \FOR{$i$ $\xleftarrow{}$ $1$ $\bold{to}$ $m$}
   \STATE $D_{t+1}(i)\xleftarrow{} \frac{D_t(i) \exp(-\alpha_t y_i h_t(x_i))}{Z_t}$   \ENDFOR
   \ENDFOR
   \STATE $f \xleftarrow{} \sum_{t=1}^T \alpha_t h_t$
   \STATE $\bold{return}$ $f$
\end{algorithmic}
\end{algorithm}

A more intuitive interpretation of AdaBoost is that the algorithm aims to combine the base classifiers by assigning particular weights to each of them. Each weight is calculated in accordance with the number of misclassifications the base classifiers return. That makes the final combined prediction of the ensemble model more robust \cite{Opitz1999}.

\subsection{Factors That Influence AdaBoost's Performance}

\textbf{The base learner} is the individual model taken from a certain family of functions $\mathcal{H}$ that is used to make predictions. For the experiment, we chose a $d-1$ dimensional perceptron via \verb+sklearn.Perceptron+ due to its VC-dimension \cite{vapnik95} being equal to $d$, as required by equation (\ref{main}). In this paper, we will refer to the set of base learners as a vector $\mathbf{h}$.

\textbf{The weight coefficients} are the real numbers that represent the significance of each individual prediction of a base learner in the final ensemble. We will denote weights as a single vector $\boldsymbol\alpha$.

\textbf{The number of iterations} determines the number of base learners that are used. It has been generally observed that the more base learners are used, the better the performance of the model. Surprisingly, the number of rounds of boosting (referred to as $T$) does not appear in the generalization bound.

\textbf{The data set} used for training also has a significant impact on the performance of AdaBoost, as boosting is particularly effective on datasets with a large number of features \cite{Opitz1999}.

\subsection{Geometric Margin Over a Dataset}

\textbf{The $L_1$-geometric margin} $\rho_f$ of a linear function $f = \sum_{t=1}^{T}\alpha_t h_t$ over a dataset $S = (x_1, \dots, x_n)$, is defined as,
\[
\rho_f = \min_{i \in [m]} \frac{|\boldsymbol\alpha \cdot \mathbf{h}(x_i)|}{\lVert \boldsymbol\alpha \rVert_1} = \min_{i \in [m]} \frac{\left|\sum_{t=1}^{T}\alpha_t h_t(x_i)\right|}{\sum_{t=1}^{T} |\alpha_t|}.
\]
The margin serves an important role in error-bound analysis, as it indicates the ``separability`` of classes. That is, the larger the margin, the more separable the clusters in the dataset are for a function $f$, and the easier the classification task will be.

\subsection{Ensemble VC-Dimension Margin Bound}

In \cite{foundations_ml}, there is the following error bound:

\textbf{Theorem.} Let $\mathcal{H}$ be a family of functions taking values in $\{+1, -1\}$ with VC-dimension $d$. Select a sample set $S$ with size $m$ and fix $L_1$-geometric margin $\rho$. Then, for any $\delta > 0$, with probability at least $1 - \delta$, the following holds for all $h \in \mathrm{conv}(\mathcal{H})$
\begin{equation}\label{main}
    R(h) \le \hat{R}_{S, \rho}(h) + \frac{2}{\rho} \sqrt{\frac{2d \log \frac{em}{d}}{m}} + \sqrt{\frac{\log\frac{1}{\delta}}{2m}},
\end{equation}
where $e$ is the Euler's constant, $\mathrm{conv}(\mathcal{H})$ is the convex hull of $\mathcal{H}$, $R(h)$ is the true error, and $\hat{R}_{S, \rho}$ is the training error (misclassification rate).

\section{Methodology}

The error of AdaBoost will be analyzed through experimental data, which will come from randomly generated datasets and the \href{https://www.kaggle.com/datasets/alexteboul/heart-disease-health-indicators-dataset}{``Heart Disease Health Indicators`` dataset} with varied properties such as size and dimensionality. For synthetic data, the \verb+sklearn.make_classification+ with parameters \verb+class_sep=0.5+ and \verb+flip_y=0.05+ will be used to generate two Gaussian clusters for binary classification. Each dataset will be split into equal-in-size training and testing sets via \verb+sklearn.train_test_split+. The train set will be used to fit the AdaBoost classifier, and the misclassification rates for both sets will be recorded.

We will conduct three experiments, investigating the influence of the sample size of the train set $m$, VC-dimension of the base learner $d$, and the number of AdaBoost's iteration $T$ on the difference of the training and testing errors, which will be denoted as $\Delta R(h) := R(h) - \hat{R}_{S, \rho}(h)$. Then, we will evaluate the theoretical error bound from the equation (\ref{main}) and look at the relationship between $\Delta R$ and $m$, $d$, $T$.

In the following experiments, we will fix the parameter $\delta$ to be equal to 0.05.

\section{Experimental Results}

\subsection{Effect of the Number of Iterations}

\begin{figure}[h]
% \vskip 0.0in
\begin{center}
\centerline{\includegraphics[width=\columnwidth]{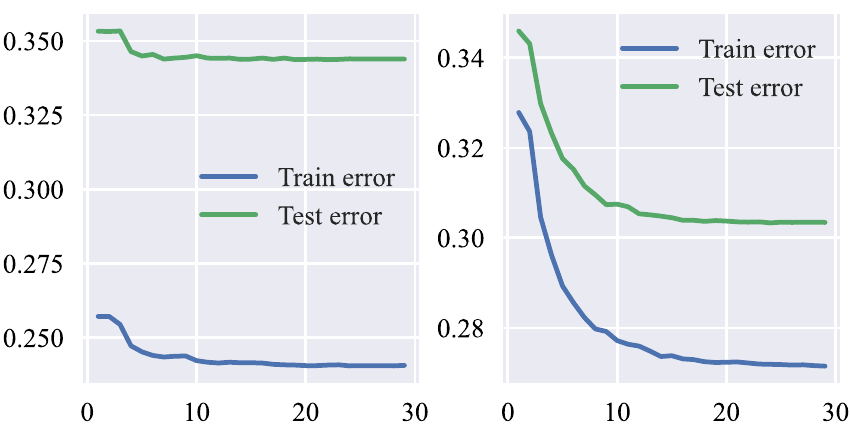}}
\caption{Results of the experiment 3.1 for $d=50, m=1000$ (left) and $d=100, m=500$ (right). The $x$-axis represents the number of iterations $T$, while the blue and green lines on the $y$-axis represent the errors on the training and testing sets, respectively.}
\label{experiment_T}
\end{center}
\vskip -0.2in
\end{figure}

First, we will test the influence of the number of base learners $T$ on the error. We ran two experiments with different parameters for $d$ and $m$, evaluated the train/test errors of the classifier, and averaged them by 100 iterations. The results are shown in Figure \ref{experiment_T}.

As can be seen in the graph, the test error looks like the train error but shifted up by a constant amount. Hence the difference between errors is also approximately constant, meaning that $\Delta R$ is not affected by $T$.

\subsection{Effect of the Sample Size}

The equation (\ref{main}) can be rewritten as
\begin{equation}\label{suppl}
\Delta R(h) \le \frac{2}{\rho} \sqrt{\frac{2d \log \frac{em}{d}}{m}} + \sqrt{\frac{\log\frac{1}{\delta}}{2m}}.
\end{equation}

Denote the right hand side as $\epsilon_{\mathrm{boost}}(\rho, d, m, \delta)$. The inequality above suggests that $\Delta R(h) = O\left(\sqrt{\frac{\log m}{m}}\right)$ and the difference of errors will slowly decrease without exceeding the theoretical bound.

We will verify this hypothesis by the following steps:
\begin{enumerate}[topsep=0pt,itemsep=-1ex,partopsep=1ex,parsep=1ex]
    \item Choose $d$ to be equal to 25, 50, 75, and 100.
    \item Generate train and test sets with dimension $d-1$, the $L_1$-margin $\rho$, and varying sample size $m$ from 10 to 10000 with step 10.
    \item Calculate the theoretical error bound $\epsilon_{\mathrm{boost}}(\rho, d, m, \delta)$.
    \item Find the difference of error on train and test sets $\Delta R(h)$.
    \item Scatter plot $\Delta R$ versus $\epsilon_{\mathrm{boost}}$.
\end{enumerate}

\begin{figure}[h!]
% \vskip 0.0in
\begin{center}
\centerline{\includegraphics[width=\columnwidth]{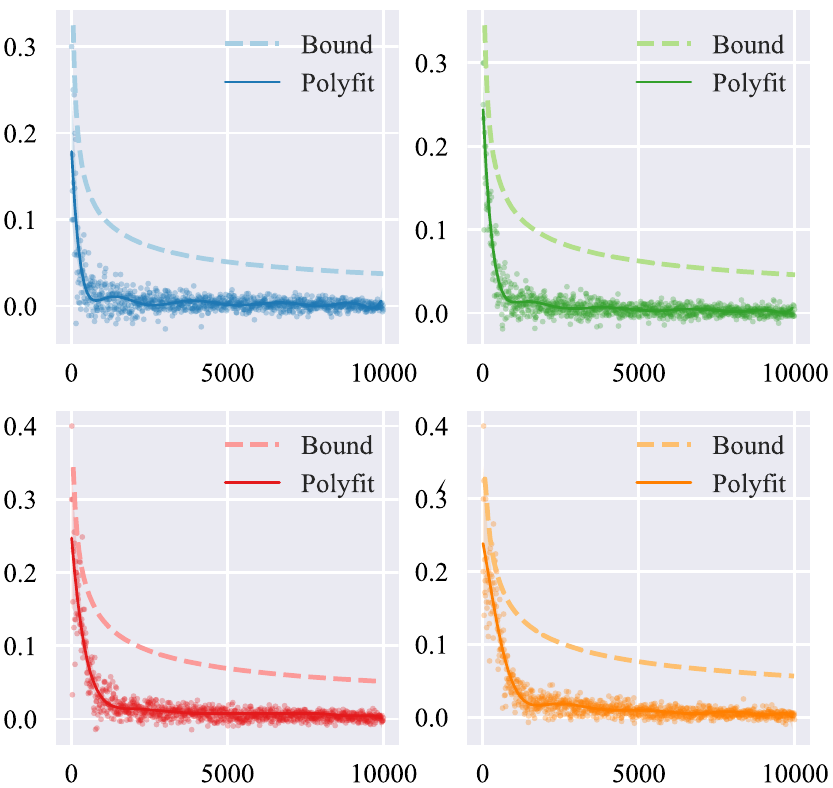}}
\caption{Results of the experiment 3.2 for different VC-dimensions $d$: the blue graph is $d=25$, the green graph is $d=50$, the red graph is $d=75$, and the yellow graph is $d=100$. The $x$-axis is the sample size $m$, the $y$-axis is the difference of the error on training and testing sets $\Delta R$. The solid line represents a polynomial fit for the training data. The dashed line represents the theoretical error bound.}
\label{experiment_m}
\end{center}
\vskip -0.2in
\end{figure}

The results can be seen in Figure \ref{experiment_m}. For clarity, we provided a polynomial fit of order 10. As expected, $\Delta R$ doesn't exceed the error bound and stays around 0 as we increase $m$.

\subsection{Effect of the Base Learners' VC-dimension}

Analogously, we can derive $\Delta R(h) = O\left(\sqrt{Cd - d\log d}\right)$ ($C$ is large enough number) from the equation (\ref{suppl}). It suggests that the difference between errors will increase quickly up to a certain point, then decrease slowly after that, also without exceeding the theoretical bound.

We will verify that by the similar steps:
\begin{enumerate}[topsep=0pt,itemsep=-1ex,partopsep=1ex,parsep=1ex]
    \item Choose $m$ to be equal to 500, 1000, 1500, and 2000.
    \item Generate train and test sets with sample size $m$, the $L_1$-margin $\rho$ and varying dimension $d$ from 5 to 1000.
    \item Calculate the theoretical error bound $\epsilon_{\mathrm{boost}}(\rho, d, m, \delta)$.
    \item Find the difference of error on train and test sets $\Delta R(h)$.
    \item Scatter plot $\Delta R$ versus $\epsilon_{\mathrm{boost}}$.
\end{enumerate}

\begin{figure}[h!]
% \vskip 0.0in
\begin{center}
\centerline{\includegraphics[width=\columnwidth]{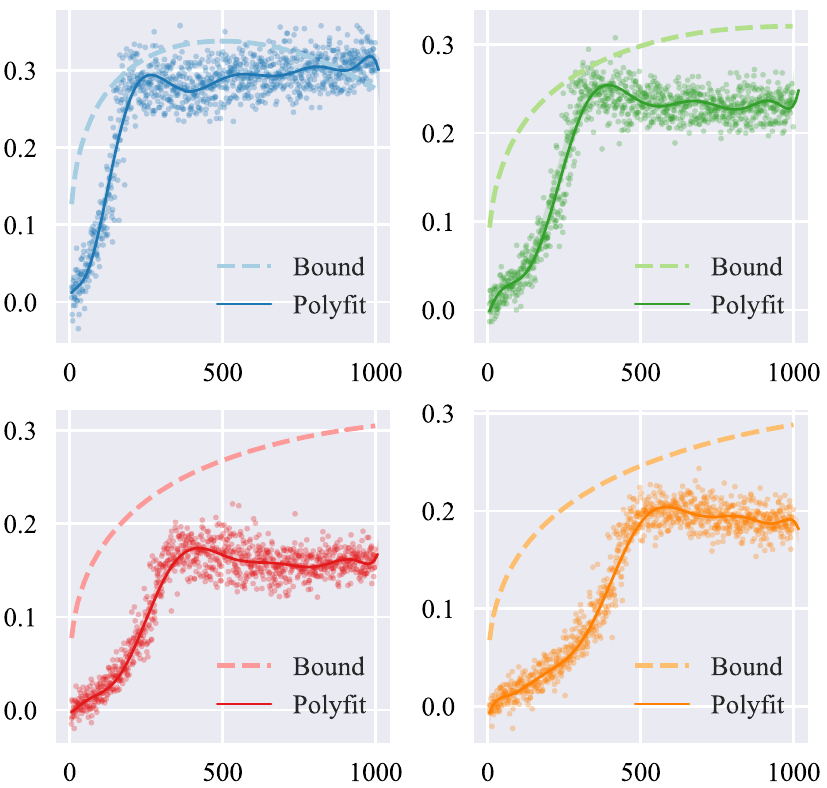}}
\caption{Results of the experiment 3.3 for different sample size $m$: the blue graph is $m=500$, the green graph is $m=1000$, the red graph is $m=1500$, and the yellow graph is $m=2000$. The $x$-axis is the VC-dimension $d$, and the $y$-axis is the difference of the error on training and testing sets $\Delta R$. The solid line represents a polynomial fit for the training data. The dashed line represents the theoretical error bound.}
\label{experiment_d}
\end{center}
\vskip -0.2in
\end{figure}

The results are shown in Figure \ref{experiment_d}. Indeed, for $m=1500$ and 2000, the difference in errors stays below the theoretical bound. However, for $m=500$ and $1000$, some values of $\Delta R$ exceed the bound.

\subsection{Evaluation of the Confidence Parameter}

Denote $(1-\delta)\cdot 100\%$ as the confidence parameter. Recall that we set $\delta = 0.05$. It means that with a $95\%$ chance, the equation (\ref{main}) will hold. Let the experimental confidence be the proportion of parameters $(m,d)$ when the equation is held from the list of all selected parameters. These experimental confidences are provided in Table 1.

\begin{table}[h]
\centering
\begin{tabular*}{\columnwidth}{@{\extracolsep{\fill}} |cc|cc|}
\hline
Exp. 3.2  & Confidence & Exp. 3.3 & Confidence \\
\hline
$d=25$    & 100\%  & $m=500$  & 82.5\% \\
$d=50$    & 99.9\% & $m=1000$ & 99.3\% \\
$d=75$    & 99.7\% & $m=1500$ & 100\%  \\
$d=100$   & 98.9\% & $m=2000$ & 100\%  \\
\hline
\end{tabular*}
\label{confid}
\centering
\caption{Experimental confidences for experiments 3.2 and 3.3 for all parameters $m$ and $d$, respectively.}
\end{table}

It is apparent that in approximately seven out of eight instances, the experimental confidence remains close to 99\%, substantially higher than 95\%. This may be due to the fact that the training data was generated from a normal distribution, thus rendering it very suitable. However, equation (\ref{main}) does not specify what the initial distribution entailed.

\subsection{Experiments on Real Data}

\begin{figure}[H]
% \vskip 0.0in
\begin{center}
\centerline{\includegraphics[width=\columnwidth]{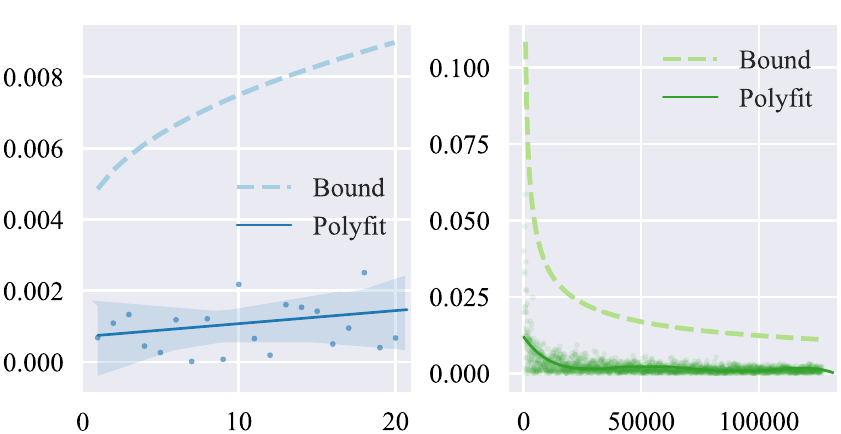}}
\caption{Results of the experiments 3.5. The left graph shows how error varies with respect to the VC-dimension of the base classifier; the $x$-axis is the VC-dimension $d$. The right graph shows the error when the sample size $m$ varies; $m$ is on the $x$-axis. The $y$-axis is the difference between the error on training and testing sets $\Delta R$ for both graphs. The solid line represents a polynomial fit for the training data. The dashed line represents the theoretical error bound.}
\label{experiment_real}
\end{center}
\vskip -0.2in
\end{figure}

We chose the ``Heart Disease Health Indicators`` dataset since it provides us with enough features and datapoints to run the proposed experiments and is designed for the binary classification tasks.

The total number of datapoints in the dataset is $253680$, and the total number of features is $22$. When running the experiments, we split the dataset into equal train and test splits, each with a total of $126840$ datapoints to simplify the experimental procedures when calculating the theoretical error bound.

In order to analyze the effect of the sample size on the error of the AdaBoost algorithm, we set the VC-dimension of the base classifiers at $21$, i.e. using all of the available features and varying $m$ (the training sample size) from $50$ to $126840$ with a step size of $50$. Similarly, when assessing the effect of the base classifier's VC-dimension, we vary the dimensionality of the inputs from $2$ to $22$ while fixing $m$ at $126840$. It is important to note that we do not vary $d$ in a random manner but feed in the features sorted by their importance. Their importance is calculated via the \verb+feature_importances_+ method. The results are summarized in Figure \ref{experiment_real}.

As we can see, the empirical error behaves as expected and does not exceed the theoretical bound for both cases anywhere on the graph.

\section{Conclusion}

In this work, we have provided an empirical verification for the error bound of the AdaBoost algorithm. As the results show, we see that the bound holds for both the synthetic and real data, which was the initial purpose of this report. 

\section{Author Contributions}

Theoretical analysis, A. B.; methodology A. B. and K. D.; synthetic data experiments, A. B.; real data experiments, K. D.; visualization, A. B.; editing, A. B. and K. D.; supervision, Zhenisbek Assylbekov. 

\bibliography{main}
\bibliographystyle{icml2022}

\end{document}